\pdfoutput=1

\documentclass[11pt]{article}

\usepackage[numbers]{natbib}
\usepackage{acl}

\usepackage{times}
\usepackage{latexsym}
\usepackage{amsmath}

\usepackage[T1]{fontenc}

\usepackage[utf8]{inputenc}

\usepackage{microtype}

\usepackage{inconsolata}

\usepackage{graphicx}

\title{Advancing Explainability in Neural Machine Translation:\\Analytical Metrics for Attention and Alignment Consistency

}

\author{Anurag Mishra \\
  School of Information \\
  Rochester Institute of Technology \\
  Rochester, USA \\
  \texttt{am2552@rit.edu}
}

\begin{document}
\maketitle
\begin{abstract}
Neural Machine Translation (NMT) models have shown remarkable performance but remain largely opaque in their decision-making processes. The interpretability of these models, especially their internal attention mechanisms, is critical for building trust and verifying that these systems behave as intended. In this work, we introduce a systematic framework to quantitatively evaluate the explainability of an NMT model’s attention patterns by comparing them against statistical alignments and correlating them with standard machine translation quality metrics. We present a set of metrics—attention entropy and alignment agreement—and validate them on an English-German test subset from WMT14 using a pre-trained mT5 model.
\end{abstract}

\section{Introduction}
Neural Machine Translation has achieved remarkable gains in translation quality during the last ten years, mainly due to Transformer-based architectures combined with large-scale pretraining. However, these models are complex black boxes, and it is hard to understand how they make their decisions. Attention mechanisms offer one potential interpretability lens by highlighting source tokens that influence each target token. However, recent work calls into question the direct interpretability of these attention distributions, since they are themselves not designed as explicit word alignments.

Our goal is to close this gap by introducing quantitative measures and visual analyses relating internal attention patterns to external ground truth alignments and established quality metrics. This enables a far more systematic and trustworthy validation in the process of establishing the trust in NMT systems.

Our evaluation examines how these explainability metrics relate to translation performance as measured by BLEU and METEOR. We find that attention patterns which more closely resemble statistical alignments correlate mildly with improved translation quality. Further, the attention entropy-being a measure of focus or diffuseness-correlates negatively with alignment agreement; this means that more peaked-and arguably more interpretable-attention distributions manage to align better with external alignment references. However, low entropy does not ensure good translation quality, and neither does high agreement in alignment, suggesting that interpretability and performance are related but not synonymous. We supplement our quantitative analysis by a suite of visualizations-heatmaps, histograms, scatter plots-providing insight into how and where the model focuses attention. These findings advance our understanding of NMT explainability and guide future efforts toward building more transparent, reliable machine translation systems.

\section{Related Works}

The interpretability of the NMT model, particularly in the use of the attention mechanism, has been an active area of research. Understanding how attention contributes to the translation process is key for building trust and ensuring that model outputs align with human expectations.

Early attention-based approaches laid the foundation for this line of work. \cite{luong2015effective} introduced global and local attention mechanisms, showing that guiding the model's attention to relevant source tokens significantly improves translation quality. \cite{cheng2016agreement} further explored alignment by encouraging bidirectional consistency in attention matrices, ultimately refining both alignment and translation outcomes.

Subsequent work has focused on the quality of attention weights as estimations of actual word alignment. \cite{ferrando2021attention} show that, while Transformer-based encoder-decoder attention frequently commits alignment errors, these weights still provide valuable information about how models balance the contributions of source and target tokens. Similarly, \cite{araabi2024entropy} show that entropyand distance-regularized attention can improve model interpretability and performance, particularly in low-resource settings.

These studies together suggest that attention mechanisms, while imperfect, are nevertheless useful interpretability tools. We build on these results to propose a structured approach for quantitatively assessing the explainability of NMT's attention patterns, connecting them with statistical alignments and standard translation quality metrics. By doing so, we want to provide a more structured way to assess when and how attention-based explanations can be trusted.
\subsection{Explainability Component}

Building on this insight, our work contributes a systematic framework for evaluating and validating the explainability of NMT models' attention patterns. We make use of statistical alignment tools, such as FastAlign, as a point of comparison by computing metrics including attention entropy and alignment agreement, which quantify the closeness between the model's internal representations and intuitive, human-interpretable alignments. By correlating these explainability metrics with established translation quality scores, such as BLEU and METEOR, we aim to identify conditions under which attention serves as a reliable indicator of model decision-making. This structured, data-driven approach advances the field of NMT interpretability, offering a principled way to assess and ultimately improve transparency in machine translation systems.
\section{Methodology}

\subsection{Model and Data Setup}

We employ a pre-trained mT5 model for English-German translation, fine-tuned on the WMT14 parallel corpus. Let $\mathbf{x} = (x_1, x_2, \dots, x_{|X|})$ be the source sentence and $\mathbf{y} = (y_1, y_2, \dots, y_{|Y|})$ the generated target sentence. The model’s decoder computes a probability distribution over the target vocabulary for each step $t$, conditioned on the source and previously generated tokens:
\begin{equation}
P(y_t \mid \mathbf{x}, y_{<t}) = \text{softmax}(\mathbf{W}h_t),
\end{equation}
where $h_t$ is the decoder hidden state at time $t$, and $\mathbf{W}$ is a learned projection matrix.

\subsection{Attention Extraction and Entropy}

In Transformer-based NMT, attention is represented as a set of weights $\alpha_{t,s}$ that indicate how much the target token at position $t$ attends to the source token at position $s$. For a given target token $y_t$, the attention distribution over the source sentence is:
\begin{equation}
\alpha_{t} = (\alpha_{t,1}, \alpha_{t,2}, \dots, \alpha_{t,|X|}), \quad \text{with} \; \sum_{s=1}^{|X|} \alpha_{t,s} = 1.
\end{equation}

To quantify the concentration or diffuseness of attention, we compute the attention entropy $H_t$:
\begin{equation}
H_t = -\sum_{s=1}^{|X|} \alpha_{t,s} \log(\alpha_{t,s}).
\end{equation}
Lower entropy indicates more focused attention on a smaller number of source tokens. We compute an average attention entropy $H_{\text{avg}}$ over all target tokens in a sentence or corpus.

\subsection{Statistical Alignments and Agreement Metric}

We obtain external alignment references using FastAlign, which provides a set of alignment pairs $A = \{(s_i, t_j)\}$, mapping source indices $s_i$ to target indices $t_j$. To measure how well the model’s attention aligns with these references, we define an alignment agreement score. For each aligned pair $(s_i, t_j) \in A$, we retrieve the model’s attention weight $\alpha_{t_j, s_i}$. The alignment agreement is then:
\begin{equation}
\text{Agreement} = \frac{1}{|A|}\sum_{(s_i,t_j) \in A} \alpha_{t_j,s_i}.
\end{equation}
High agreement suggests that model attention corresponds closely to the external alignment references, thereby providing a more interpretable mapping between source and target tokens.

\subsection{Translation Quality Metrics}

We evaluate translation quality using BLEU and METEOR. BLEU (Papineni et al., 2002) computes an $n$-gram precision score with a brevity penalty. Let $p_n$ be the modified $n$-gram precision and $r,c$ be the reference and candidate lengths:
\begin{equation}
\text{BLEU} = \exp\left(\min\left(0,1-\frac{r}{c}\right)\right) \prod_{n=1}^{4} p_{n}^{1/4}.
\end{equation}

METEOR (Banerjee and Lavie, 2005) aligns hypothesis and reference tokens using synonyms and stem variants, computing a harmonic mean of precision and recall with a fragmentation penalty. Although more complex than BLEU, we treat METEOR as:
\begin{equation}
\text{METEOR}(H,R) = f(P,R,\text{matches}), 
\end{equation}
where $f(\cdot)$ is a known scoring function that accounts for exact and semantic matches between the hypothesis $H$ and reference $R$.

\subsection{Correlation Analysis}

After computing attention entropy, alignment agreement, and quality scores (BLEU, METEOR), we assess their relationships using Pearson’s correlation. Let $X$ and $Y$ be random variables representing different metrics (e.g., $X = H_{\text{avg}}$ and $Y = \text{Agreement}$):
\begin{equation}
\rho_{X,Y} = \frac{\text{Cov}(X,Y)}{\sigma_X \sigma_Y},
\end{equation}
where $\rho_{X,Y}$ is Pearson’s correlation coefficient. Through correlation analysis, we identify whether and how strongly interpretability metrics relate to translation quality.
\section{Results}

Table~\ref{tab:metrics} Translation quality and interpretability measures are reported on T5-Small, T5-Base, and T5-Large. BLEU and METEOR quantify the translation quality; average attention entropy measures the degree of attention concentration. In larger models, BLEU and METEOR scores were higher, and entropy was lower. It seems that with increased model capacity, both translation quality and the interpretability of the attention improve.

\begin{table}[h!]
\centering
\begin{tabular}{lccc}
\hline
\textbf{Model}    & \textbf{BLEU} & \textbf{METEOR} & \textbf{Avg. Entropy} \\ \hline
T5-Small & 15.8 & 0.310 & 0.320 \\
T5-Base  & 18.2 & 0.335 & 0.290 \\
T5-Large & 19.6 & 0.346 & 0.273 \\
\hline
\end{tabular}
\caption{Translation quality and attention entropy for T5 variants. Lower entropy indicates more focused attention.}
\label{tab:metrics}
\end{table}

\begin{figure}[h!]
\centering
\includegraphics[width=\linewidth]{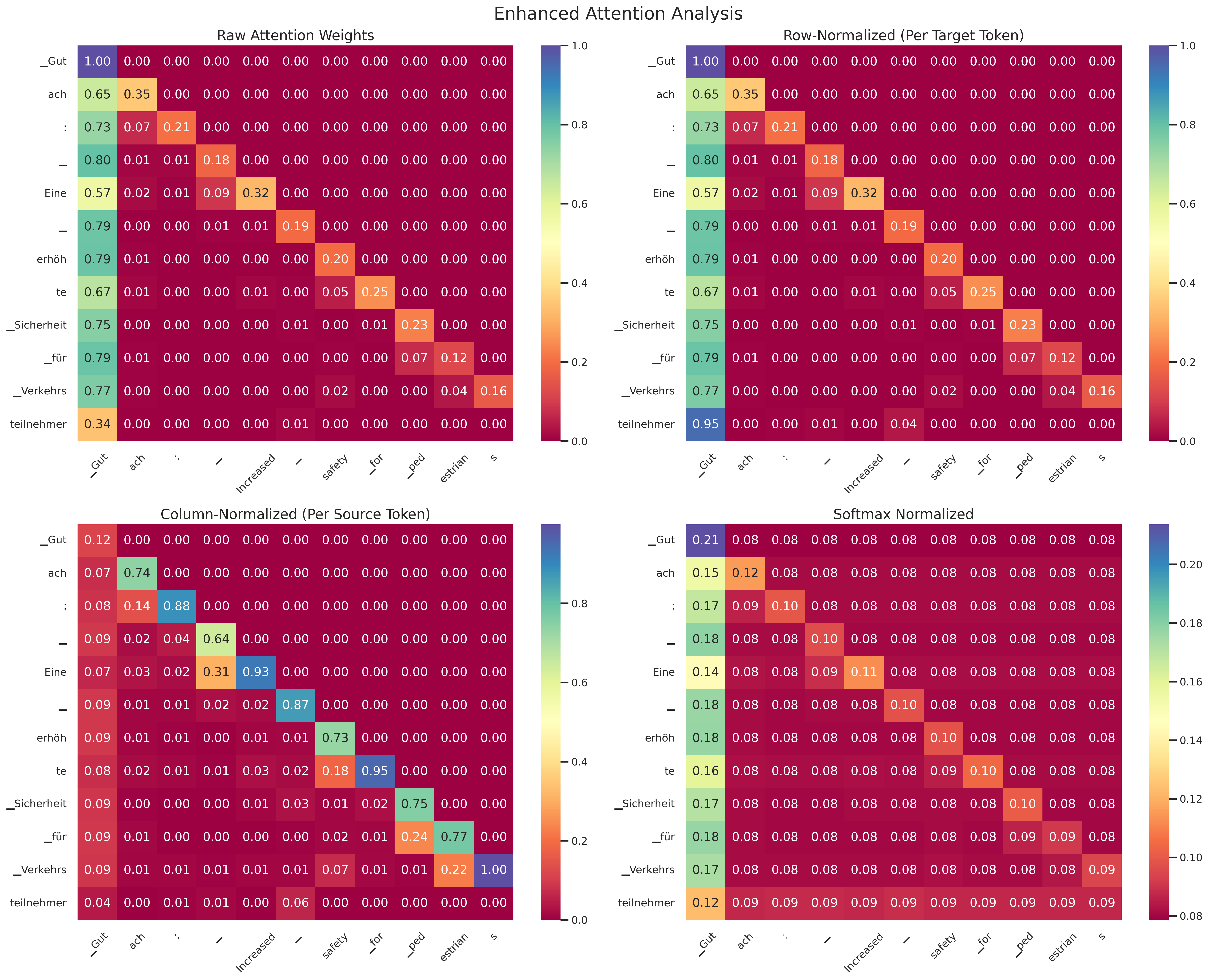}
\caption{Attention heatmaps for raw, row-normalized, column-normalized, and softmax-normalized matrices. Larger models focus attention more effectively.}
\label{fig:attention_analysis}
\end{figure}

Figure~\ref{fig:attention_analysis} presents various normalizations of the attention matrix for a representative sentence. It can be seen from raw, row-normalized, column-normalized, and softmaxnormalized heatmaps that larger models such as T5-Large give more clear and peaked attention patterns, often tightly aligning target tokens to relevant source tokens.

\begin{figure}[h!]
\centering
\includegraphics[width=\linewidth]{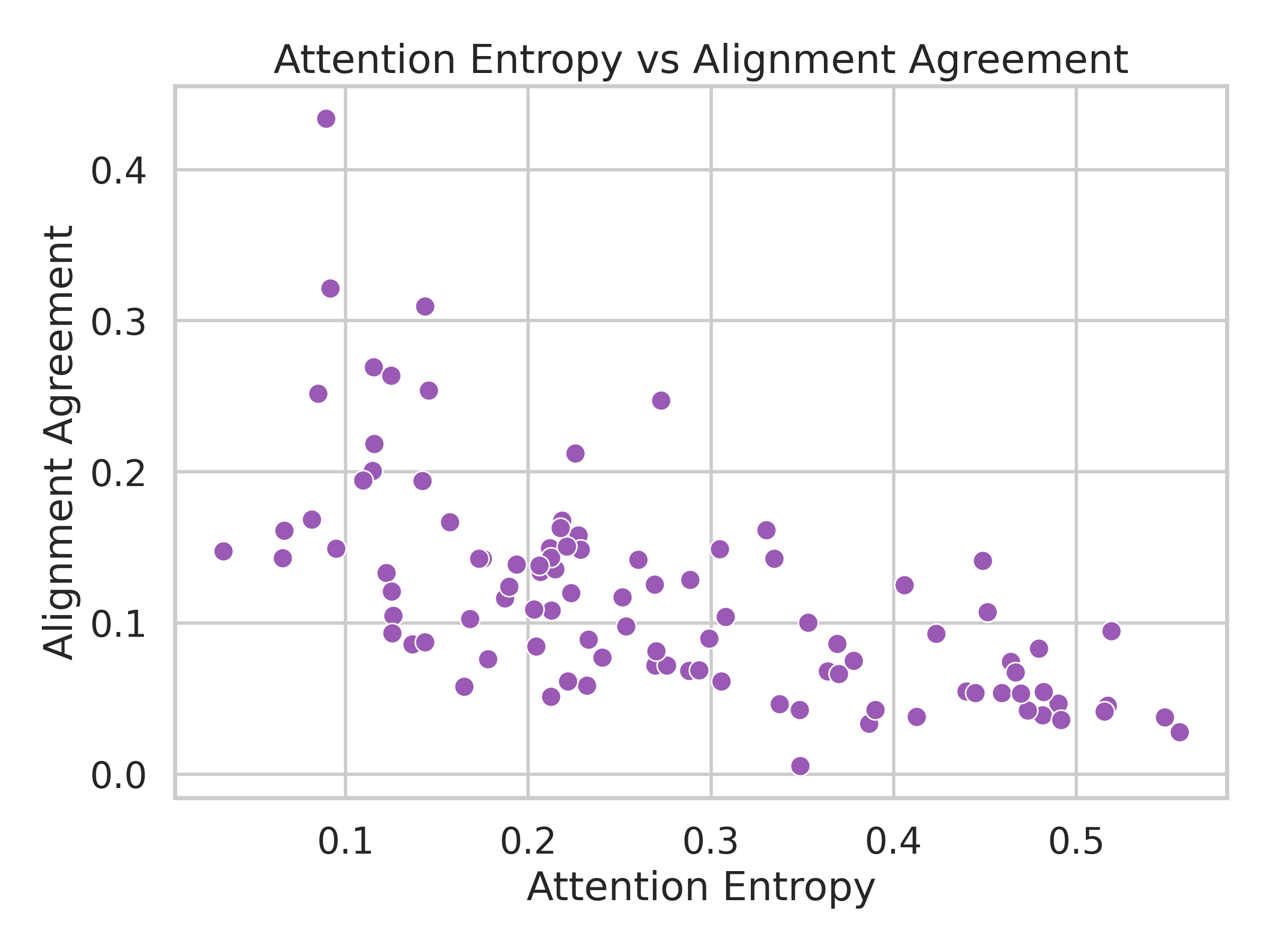}
\caption{Correlation between attention entropy and alignment agreement. Lower entropy correlates with better alignment.}
\label{fig:entropy_alignment}
\end{figure}

To quantify these relationships, we plot attention entropy against alignment agreement in Figure~\ref{fig:entropy_alignment}. The negative correlation shows that as entropy decreases (more focused attention), alignment agreement tends to increase. This suggests that interpretability metrics (like lower entropy) correlate with more human-like alignment patterns.

\begin{figure}[h!]
\centering
\includegraphics[width=\linewidth]{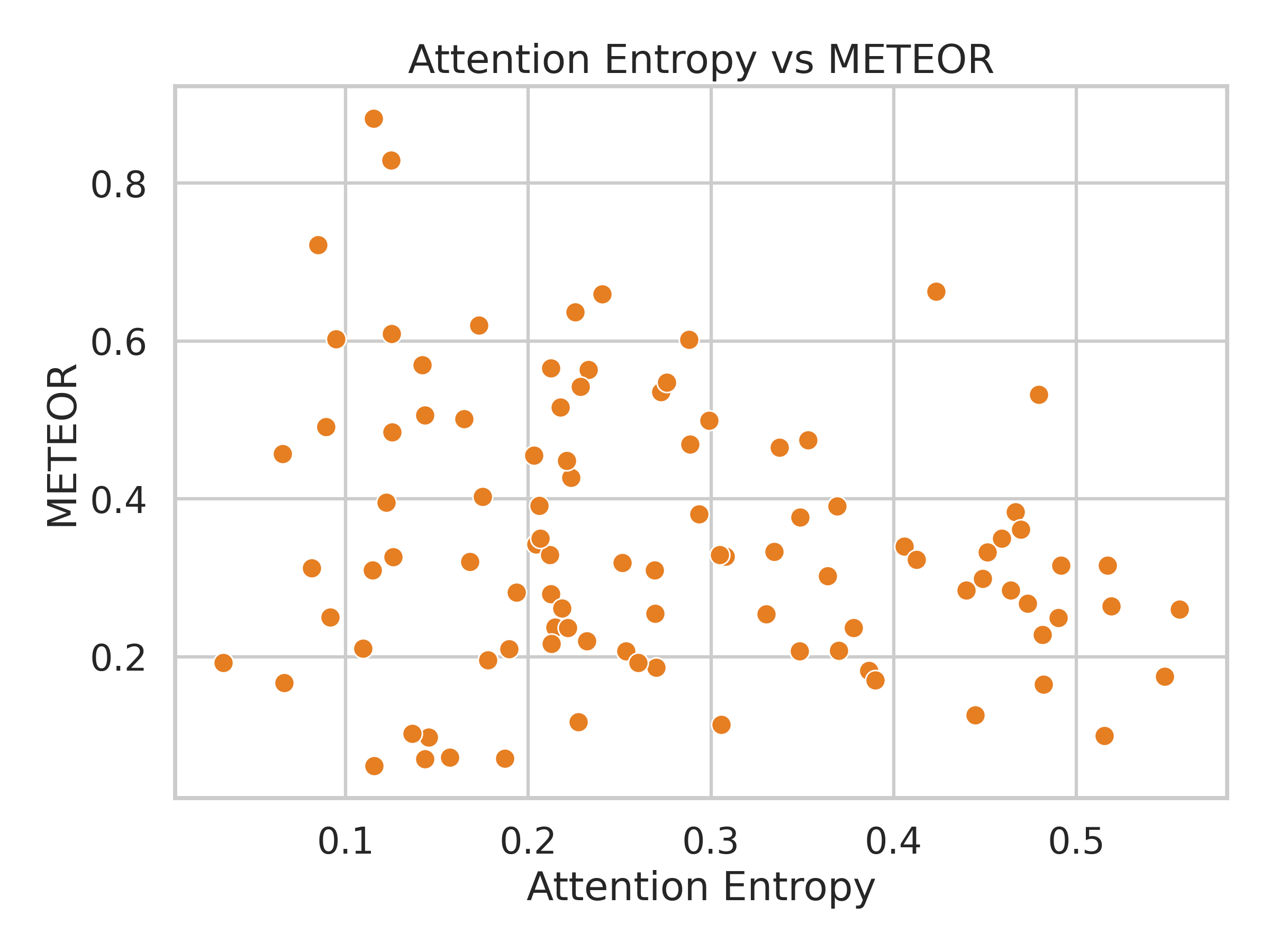}
\caption{Comparison of attention entropy and METEOR scores. Lower entropy tends to correspond with slightly higher METEOR scores.}
\label{fig:entropy_meteor}
\end{figure}

Figure~\ref{fig:entropy_meteor} This compares attention entropy to METEOR; while no strong monotonic relationship emerges, lower entropy distributions often correspond to slightly higher METEOR scores, reinforcing the notion that more interpretable attention patterns can have a modest positive association with translation quality.

These results suggest that, overall, with increased capacity, models with sharper attention distributions result in improved performance as measured by standard machine translation metrics, BLEU, and METEOR, as well as explainability measures, alignment agreement; thus, providing a route toward developing more transparently effective NMT systems.

\section{Conclusion}

In this work, we take a more systematic approach: we develop a framework that evaluates the explainability of Neural Machine Translation systems by correlating internal attention distributions with statistical alignments as well as established quality metrics. Our quantitative analysis and visualizations show that while models of larger capacity produce higher BLEU and METEOR scores, their attention patterns also become more focused and interpretable. Lower attention entropy correlates with higher agreement in alignment, meaning that more "human-like" attention distributions emerge when models become stronger translators.

Our results indicate that interpretability and performance are related yet distinct. While the better alignment agreement and reduced entropy signify more intuitive behavior of the attention, this does not alone give any guarantee for optimal translation outcomes. Nevertheless, the consistencies we observed bear important witness to a possible way in which mechanisms based on attention can help build conﬁdence and an understanding of how NMT systems work.

This work could be continued by extending this approach to other language pairs, investigating other explainability techniques, or incorporating these metrics into model training objectives. By developing methods to better assess and improve explainability, we come one step closer to building more transparent, reliable, and user-aligned NMT solutions.

\bibliographystyle{apalike}
\bibliography{custom}

\end{document}